\documentclass[final]{cvpr}

\usepackage{times}
\usepackage{epsfig}
\usepackage{graphicx}
\usepackage{amsmath}
\usepackage{amssymb}
\usepackage{booktabs,amsfonts}
\usepackage[noend]{algpseudocode}
\usepackage{algorithmicx,algorithm}
\usepackage{makecell}
\usepackage{setspace}

\usepackage[pagebackref=true,breaklinks=true,colorlinks,bookmarks=false]{hyperref}

\def\cvprPaperID{3362} 
\def\confYear{CVPR 2021}

\begin{document}

\title{A Circular-Structured Representation for Visual Emotion Distribution Learning}


\author{Jingyuan~Yang{\textsuperscript{1}}
	\qquad Jie~Li{\textsuperscript{1}}
	\qquad Leida~Li{\textsuperscript{2}}
	\qquad Xiumei~Wang{\textsuperscript{1}}
	\qquad Xinbo~Gao{\textsuperscript{1,3}\thanks{Corresponding author}}\\
	\textsuperscript{1} School of Electronic Engineering, Xidian University, Xi'an, China \\
	\textsuperscript{2} School of Artificial Intelligence, Xidian University, Xi'an, China \\
	\textsuperscript{3} The Chongqing Key Laboratory of Image Cognition, \\ Chongqing University of Posts and Telecommunications, Chongqing, China \\	
	{\tt\small jingyuanyang@stu.xidian.edu.cn, leejie@mail.xidian.edu.cn, ldli@xidian.edu.cn,} \\ 
	{\tt\small  wangxm@xidian.edu.cn, xbgao@mail.xidian.edu.cn}
}
%

\maketitle
\pagestyle{empty}  
\thispagestyle{empty} 

\begin{abstract}
   Visual Emotion Analysis (VEA) has attracted increasing attention recently with the prevalence of sharing images on social networks. 
   Since human emotions are ambiguous and subjective, it is more reasonable to address VEA in a label distribution learning (LDL) paradigm rather than a single-label classification task. 
   Different from other LDL tasks, there exist intrinsic relationships between emotions and unique characteristics within them, as demonstrated in psychological theories.
   Inspired by this, we propose a well-grounded circular-structured representation to utilize the prior knowledge for visual emotion distribution learning.
   To be specific, we first construct an Emotion Circle to unify any emotional state within it.
   On the proposed Emotion Circle, each emotion distribution is represented with an emotion vector, which is defined with three attributes (i.e., emotion polarity, emotion type, emotion intensity) as well as two properties (i.e., similarity, additivity). 
   Besides, we design a novel Progressive Circular (PC) loss to penalize the dissimilarities between predicted emotion vector and labeled one in a coarse-to-fine manner, which further boosts the learning process in an emotion-specific way. 
   Extensive experiments and comparisons are conducted on public visual emotion distribution datasets, and the results demonstrate that the proposed method outperforms the state-of-the-art methods.
\end{abstract}

\section{Introduction}
\label{sec:introduction}
Emotion serves as one of the most essential factors that distinguish human beings from other species, which affects almost every aspect of our daily lives.
With the prevalence of social networks, more and more people tend to express their feelings by sharing images on the internet.
Therefore, \textit{Visual Emotion Analysis (VEA)} has drawn great attention recently, which aims at understanding how people feel emotionally towards different visual stimuli.
The development in VEA may also bring benefit to other related tasks (\eg aesthetic assessment~\cite{chen2020adaptive, hosu2019effective}, memorability estimation~\cite{fajtl2018amnet, sidorov2019changing}, and stylized image captioning~\cite{chen2018factual,guo2019mscap}), as well as potential applications (\eg opinion mining~\cite{li2019survey, sobkowicz2012opinion}, intelligent advertising~\cite{holbrook1984role, mitchell1986effect},  and mental disease treatment~\cite{jiang2017learning, wieser2012reduced}).

\begin{figure}
	\centering
	\includegraphics[width=\linewidth]{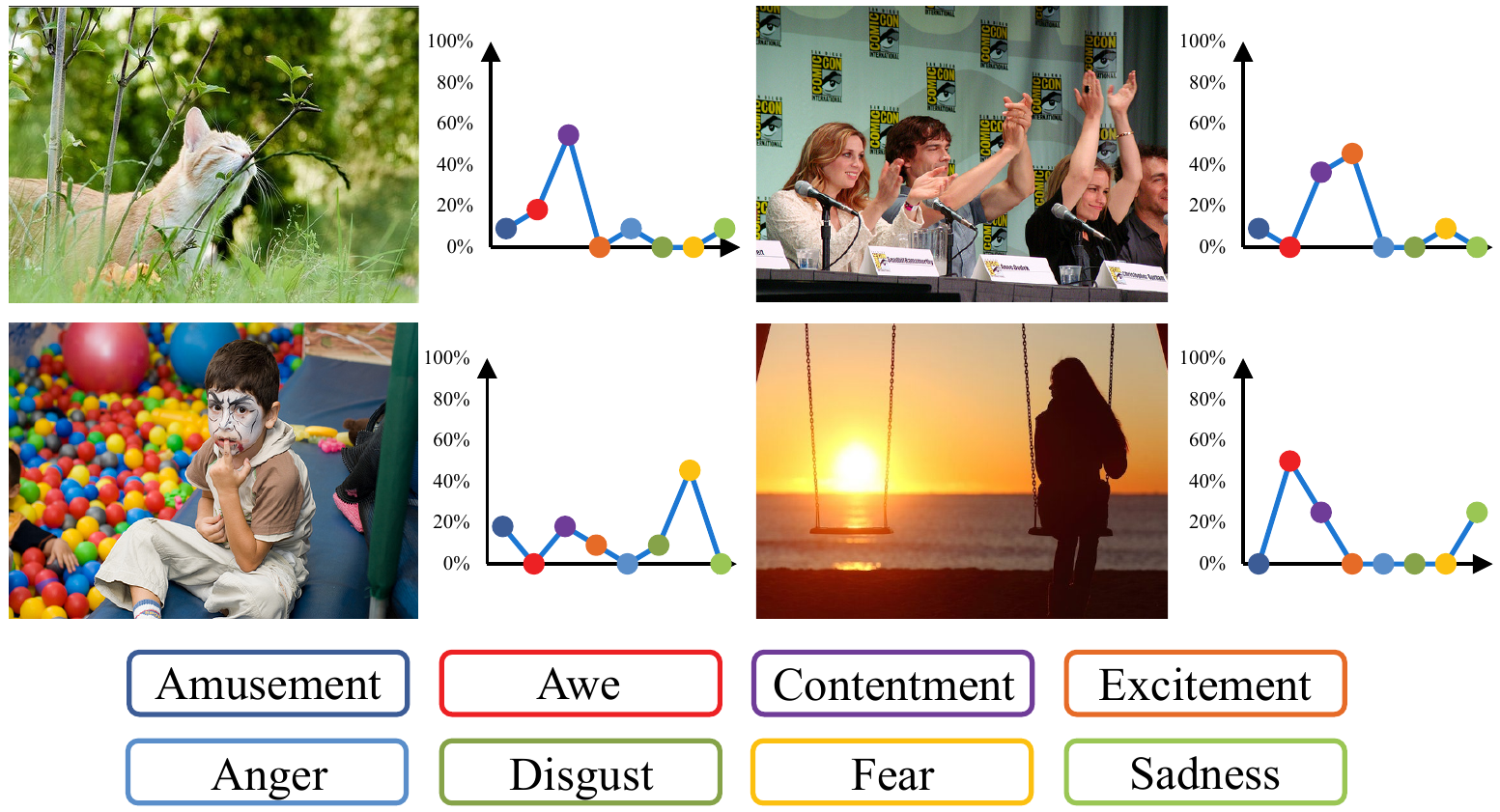}
	\vspace{-8pt}
	\caption{Four images with different emotion distributions from the involved datasets. Rather than a dominant emotion, images often evoke multiple emotions with different description degrees.
	}
	\vspace{-10pt}
	\label{fig:introduction_img}
\end{figure}

Most of the existing work often regard VEA as a single-label classification task~\cite{rao2016learning, yang2018weakly, yang2018visual, zhang2019exploring, zhu2017dependency}, assuming that each image only evokes a dominant emotion.
However, these attempts over-simplified the complexity of human emotions and neglected the ambiguity and subjectivity lies in them.
In reality, different people may experience different emotions towards one image (\ie, subjectivity), and even an individual may have diverse emotions towards one image (\ie, ambiguity).
\textit{Label Distribution Learning (LDL)} is proposed to deal with the problem when an instance is covered by a certain number of labels with different description degrees~\cite{geng2016label}.
Thus, it is more reasonable to address VEA in an LDL paradigm rather than a single-label classification task, as shown in Fig.~\ref{fig:introduction_img}.
Various methods have been proposed to deal with visual emotion distribution learning so far, including the earlier traditional algorithms~\cite{geng2016label, yang2017learning, zhao2018discrete} and the recent deep learning ones~\cite{he2019image, peng2015mixed, xiong2019structured, yang2017joint, yang2017learning}.
Most of the aforementioned methods simply implement general LDL algorithms for VEA.
However, unlike other LDL tasks, there exist intrinsic relationships between distinct emotion labels, which can be viewed as useful prior knowledge of this task.

The study of human emotions is not only limited to computer vision field, but also being heated discussed in psychology~\cite{koole2009psychology, strongman1996psychology}, biology~\cite{mcnaughton1989biology, plutchik1994psychology}, as well as sociology~\cite{hochschild1998sociology, riis2010sociology}.
Psychologists developed various emotion models to illustrate the intrinsic relationships among emotions, including the well-known Plutchik's Wheel~\cite{plutchik1980general} and Mikel's Wheel~\cite{mikels2005emotional, zhao2016predicting}.
Besides, it has been demonstrated in psychology that people experience eight basic emotions, 
which can be expressed at different intensities and can be combined to form any emotional state~\cite{plutchik1980general}.

\begin{figure}
	\centering
	\includegraphics[width=\linewidth]{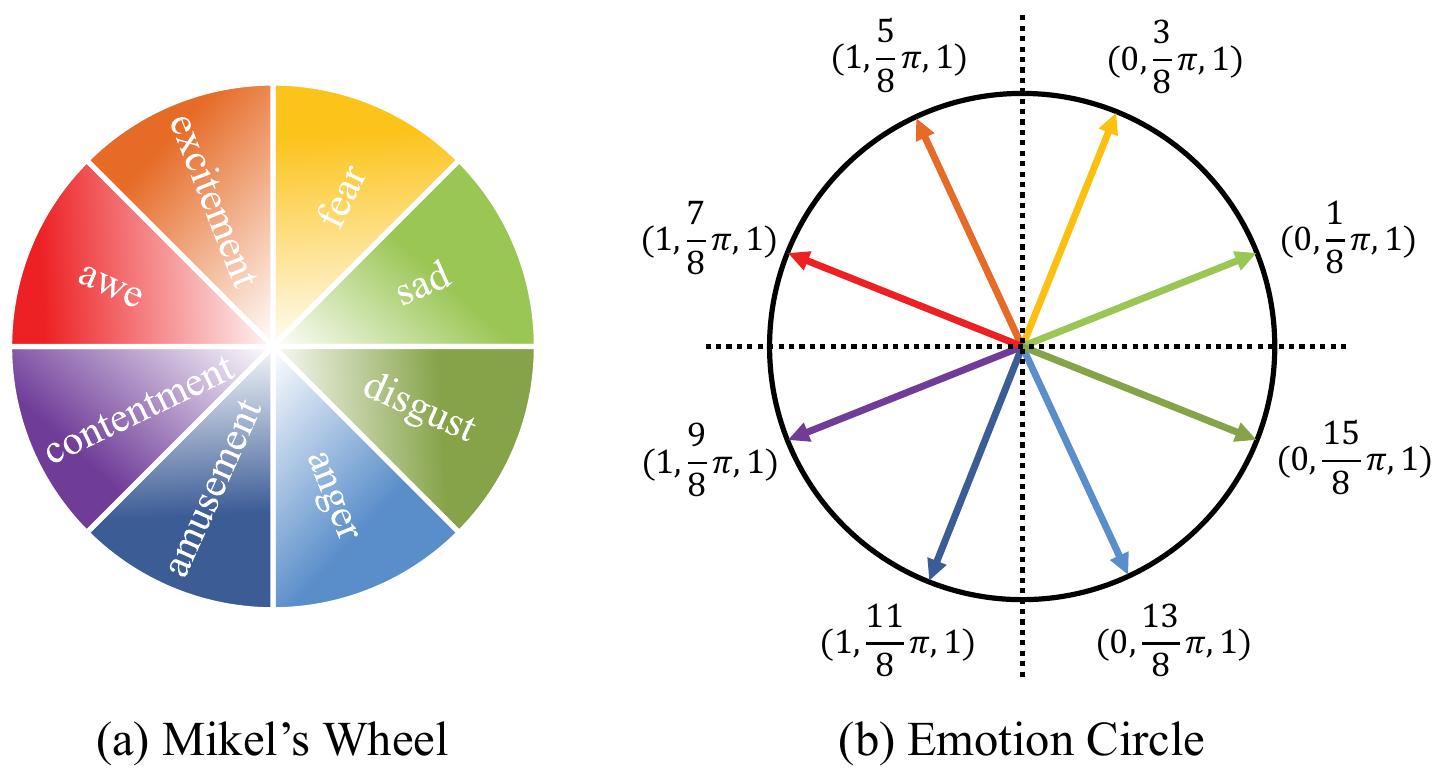}
	\vspace{-10pt}
	\caption{Mikel's Wheel from psychological model (a), and the proposed Emotion Circle (b) with eight basic emotion vectors evenly distributed in accordance with Mikel's Wheel.
	}
	\vspace{-10pt}
	\label{fig:introduction}
\end{figure}

Inspired by the above studies, we propose a well-grounded circular-structured representation for visual emotion distribution learning, aiming to effectively utilize the intrinsic relationships between emotions and unique characteristics within them.
To be specific, we construct an \textit{Emotion Circle} to unify any emotional state within it, including both basic emotions and the compound ones.
On the proposed Emotion Circle, each emotional state can be well-presented as an emotion vector with three attributes (\ie, \textit{emotion polarity}, \textit{emotion type} and \textit{emotion intensity}) and two properties, namely \textit{similarity} and \textit{additivity}.
As shown in Fig.~\ref{fig:introduction}, basic emotions are defined with a set of unit vectors evenly-distributed on the Emotion Circle (b), corresponding to the circular structure of the eight basic emotions (\ie, amusement, awe, contentment, excitement, anger, disgust, fear, sad) in Mikel's Wheel (a)~\cite{mikels2005emotional, zhao2016predicting}.
In addition to the above basic emotions, there also exist complex emotional states in reality (\eg, emotion distributions).
Therefore, considering that basic emotions have intensities and can be combined to form any emotional state~\cite{plutchik1980general}, we propose a systematic approach to map any emotion distribution to a compound emotion vector on the Emotion Circle.
In detail, each emotion distribution is first projected on basic emotion vectors with different emotion intensities according to their description degrees.
Weighted basic emotion vectors are then combined to form a compound emotion vector, which can be regarded as the specific circular-structured representation of a given emotion distribution.

Most visual emotion distribution learning methods~\cite{he2019image, yang2017joint, yang2017learning} simply leverage the \textit{Kullback-Leibler (KL) loss}~\cite{kullback1951information} to directly measure the differences between labeled emotion distributions and the predicted ones.
However, rather than a set of uncorrelated labels, emotions are closely related to each other and circularly distributed according to psychological models.
To exploit such prior knowledge, we propose a novel \textit{Progressive Circular (PC) loss} to learn the dissimilarities between the labeled emotion vector and the predicted one on the Emotion Circle in a coarse-to-fine manner.
In detail, We first construct polar loss to penalize the difference between the predicted polarity and the labeled one, which can be viewed as a coarse constraint towards a specific emotional state. 
After divide emotions into two polarities, we build up a more fine-grained constraint on emotion type by calculating the distance between them.
Since emotion intensity is regarded as a crucial factor when describing a specific emotional state~\cite{plutchik1980general}, we further leverage it as the confidence degree to constrain the above two losses.
By implementing both the KL loss and the proposed PC loss, the learning process of emotion distribution is not only optimized through its conventional mechanism, but also further boosted in a novel emotion-specific circular-structured manner.

\begin{figure*}
	\centering
	\includegraphics[width=\textwidth]{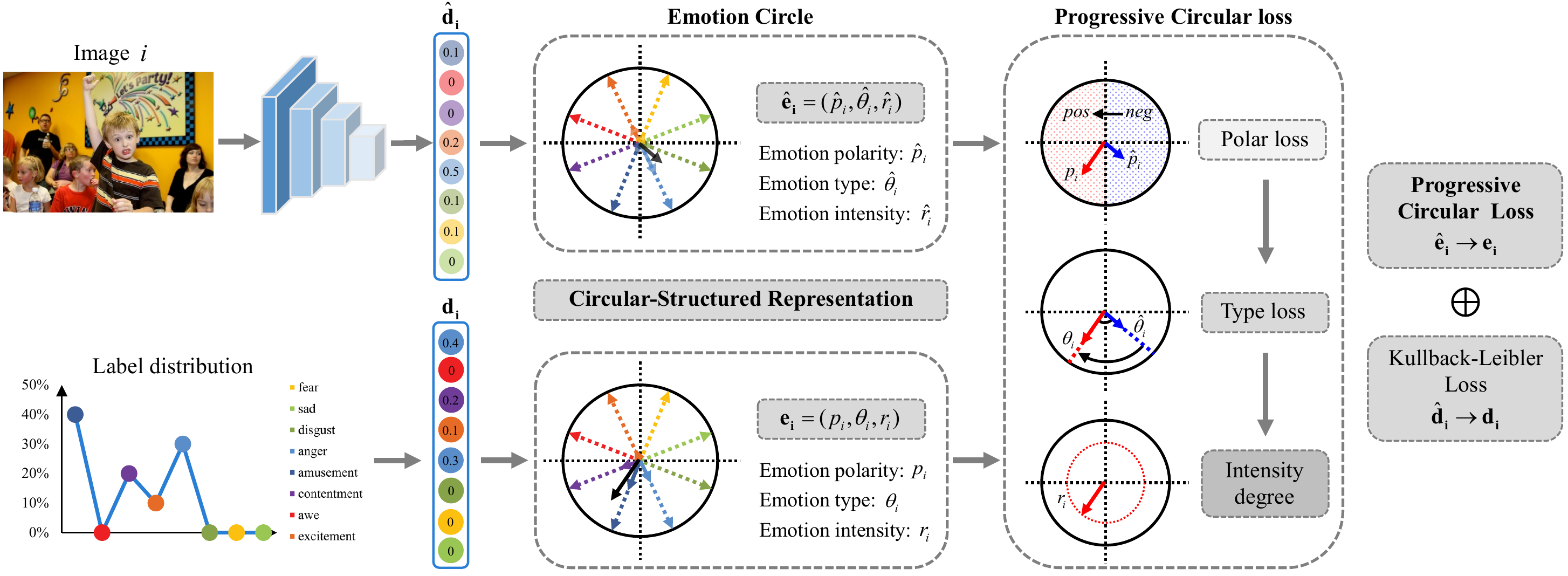}
	\vspace{-10pt}
	\caption{Framework of the proposed circular-structured representation.
		On the proposed Emotion Circle, both the predicted emotion distribution and the labeled one are represented with compound emotion vectors through a systematic approach.
		We then propose the Progressive Circular loss in a coarse-to-fine manner, which is further exploited to train the network together with Kullback-Leibler loss.
	}
	\vspace{-10pt}
	\label{fig:network}
\end{figure*}

Our contributions can be summarized as follows:
\begin{itemize}
	
	\item We propose a well-grounded circular-structured representation to learn visual emotion distribution by utilizing the intrinsic relationships between emotions, which consistently outperforms the state-of-the-art methods on several emotion distribution datasets.
	To the best of our knowledge, it is the first time to systematically exploit emotion relationships as prior knowledge for visual emotion distribution learning.
	\item We construct an Emotion Circle to unify any emotional state within it, where each emotion distribution is represented as a compound emotion vector with three attributes as well as two properties according to psychological models.
	\item We design a novel Progressive Circular loss to penalize the dissimilarities between labeled emotion vectors and the predicted ones on the Emotion Circle from coarse to fine, which further boosts the visual emotion distribution learning process in an emotion-specific circular-structured manner.
	
\end{itemize}

\section{Related Work}
\label{sec:related_work}
\subsection{Visual Emotion Analysis}
\label{sec:visual_emotion_analysis}
Researchers have been devoted to visual emotion analysis (VEA) for more than two decades~\cite{lang1997international}, during which approaches vary from the early traditional ones to the recent deep learning ones.
Earlier works in VEA mainly focused on designing hand-crafted features to mine emotions from affective images~\cite{borth2013large, machajdik2010affective, zhao2014exploring, zhao2014affective}.
Inspired by psychology and art theory, Machajdik~\etal~\cite{machajdik2010affective} extracted rich hand-crafted features including color, texture, composition, and content.
Zhao~\etal~\cite{zhao2014affective} extracted features from low-level elements-of-art, mid-level principles-of-art, to high-level semantics in a multi-graph learning framework.
Although hand-crafted features have been proven to be effective on several small-scale datasets, they are still limited to cover all the important factors in visual emotion.
Recently, researchers in VEA have adopted Convolutional Neural Network (CNN) to predict emotions and have achieved gratifying results~\cite{rao2016learning, yang2018weakly, yang2018visual, you2017visual, you2015robust, zhang2019exploring}.
A multi-level deep representation network (MldrNet) was constructed by Rao~\etal~\cite{rao2016learning} to extract emotional features from image semantics, aesthetics, and low-level features simultaneously.
Aiming to focus on local regions instead of the holistic one, Yang~\etal~\cite{yang2018visual} constructed a local branch to discover affective regions (AR) by implementing the off-the-shelf detection tools. 
A weakly supervised coupled network (WSCNet)~\cite{yang2018weakly} was further proposed by Yang~\etal to discover emotion regions as well as to predict visual emotions in an end-to-end manner.
However, these methods simply regard VEA as a single-label classification task, which neglects the ambiguity and subjectivity lies in human emotions.
To be specific, different people may evoke different emotions towards one image (\ie, subjectivity) and even an individual may experience multiple emotions towards one image (\ie, ambiguity).
Thus, rather than a single-label classification task, it is more reasonable to consider VEA in a Label Distribution Learning~\cite{plutchik1980general} paradigm.

\begin{figure*}
	\centering
	\includegraphics[width=0.8\textwidth]{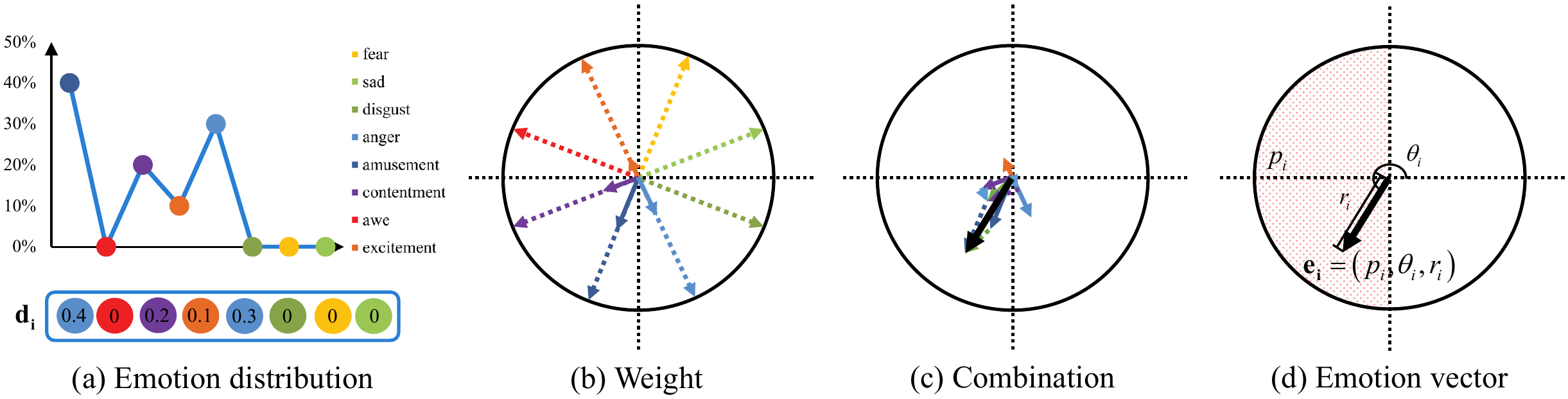}
	\caption{Mapping from the emotion distribution (a) to the compound emotion vector (d) on the Emotion Circle.
		We first weigh eight basic emotions with different description degrees (b) and then combine them to form a compound emotion vector through vector addition operations (c).
		The final emotion vector can be viewed as a specific circular-structured representation of a given emotion distribution.
	}
	\vspace{-10pt}
	\label{fig:construction}
\end{figure*}

\subsection{Label Distribution Learning}
\label{label_distribution_learning}
Label distribution learning (LDL) is proposed by Geng~\etal~\cite{geng2016label} to address the problem when an instance is covered by a certain number of labels with different description degrees, which is a more general learning framework covering both single-label and multi-label learning tasks.
Traditional algorithms in LDL can be roughly divided into three strategies, \ie, problem transfer (PT), algorithm adaption (AA), and specialized algorithms (SA)~\cite{geng2016label}.
With the help of deep networks, deep label distribution learning (DLDL) is proposed by Gao~\etal~\cite{gao2017deep} to effectively prevent the network from over-fitting by utilizing the label ambiguity in both feature learning and classifier learning.
Consequently, LDL was then introduced to VEA to deal with subjectivity and ambiguity lies in it~\cite{he2019image, peng2015mixed, xiong2019structured, yang2017joint,yang2017learning, zhao2018discrete}.
Peng~\etal~\cite{peng2015mixed} proposed convolutional neural network regressions (CNNR) to address VEA in LDL for the first time.
Subsequently, Yang~\etal~\cite{yang2017joint} proposed a joint network (JCDL) to learn visual emotion from distribution and classification simultaneously, by optimizing both softmax loss and Kullback-Leibler (KL) loss.
Most of the aforementioned methods simply employ general LDL mechanisms and consequently ignore the intrinsic relationships between emotions, which can be regarded as prior knowledge for visual emotion distribution learning.
Motivated by psychological theories~\cite{mikels2005emotional, plutchik1980general, zhao2016predicting}, we propose a circular-structured representation to utilize the intrinsic relationships between emotions and unique characteristics within it. 

\section{Methodology}
\label{sec:methodology}
\subsection{Emotion Circle}
\label{sec:emotion_circle} 
Unlike other label distribution learning tasks, emotion labels are closely related to each other with circular structure and unique characteristics based on psychological models~\cite{mikels2005emotional, plutchik1980general, zhao2016predicting}.
Inspired by this, aiming at learning emotions in a more specific and reasonable way, we construct an Emotion Circle to unify any emotional states in it.
On the proposed Emotion Circle, each emotional state is represented with an emotion vector $\mathbf{{e}_{i}}$, which is described as
\begin{align}
\label{eq:emotion_vector}
\mathbf{{e}_{i}} = (p_i,\theta_i, r_i),
\end{align}
where $p_i$, $\theta_i$, $r_i$ denotes the emotion polarity, emotion type and emotion intensity respectively.
We give detailed definitions of the above three emotion attributes as follows and further introduce two properties for emotion vectors according to psychological theories.
\vspace{-5pt}
\begin{itemize}
	\setlength{\itemsep}{0pt}
	\setlength{\parsep}{0pt}
	\setlength{\parskip}{0pt}

	\item \textbf{Emotion Polarity ($p_i$)}:
	In addition to its given emotion label (\ie, amusement, awe, contentment, excitement, anger, disgust, fear, sad), each emotion is also naturally grouped into a specific emotion polarity (\ie, positive, negative).
	Specifically, amusement, awe, contentment, excitement belong to positive emotions while anger, disgust, fear, sad are negative ones.
	Hence, we divide the Emotion Circle to half positive and half negative, as given in Eq.~\ref{eq:polar}, in accordance with the polar structure on the Mikel's Wheel~\cite{mikels2005emotional, zhao2016predicting}:
	\vspace{-10pt}
	\begin{spacing}{1.2}
	\begin{align}
	\label{eq:polar}
	{p_i}=\left\{\begin{array}{l}
	{0, \ \ \ \ \ \theta_i \in \left[ 0,\frac{1}{2}\pi \right) \cup \left[ \frac{3}{2}\pi, 2\pi \right),}    \\
	{1, \ \ \ \ \ \theta_i \in \left[ \frac{1}{2}\pi,\frac{3}{2}\pi \right).} 
	\end{array}\right.
	\end{align}
	\end{spacing}
	\vspace{-5pt}
	\item \textbf{Emotion Type ($\theta_i$)}:
	We define emotion type with the polar angle $\theta_i \in \left[ 0 , 2\pi \right]$, in order to preserve the circular-distributed emotions on the Mikel's Wheel.
	There are basic emotions and compound emotions in reality~\cite{plutchik1980general}, which both can be well-represented on the Emotion Circle.
	In Fig.~\ref{fig:introduction}(b), we first set a series of evenly-distributed unit vectors on the Emotion Circle to represent eight basic emotions, denoted as
	\begin{align}
	\label{eq:basic_emotion_theta}
	{{\theta }^{j}}=\frac{2j-1}{8}\pi, j\in \left[ 1,C \right],
	\end{align}
	where $C$ denotes the number of emotion categories in psychological models.
	Besides, there also exist compound emotion types, which are distributed in the blank space between these basic vectors.
	\item \textbf{Emotion Intensity ($r_i$)}:
	As mentioned in a psychological literature~\cite{plutchik1980general}, ``Each emotion can exist in varying degrees of intensity.''
	Thus, polar diameter $r_i \in [0,1]$ is introduced to describe the intensity of a specific emotion type $\theta_i$, where $r_i=1$ indicates the strongest degree and $r_i=0$ the weakest.
	To be specific, the emotion intensity of each basic emotion is set as
	\begin{align}
	\label{eq:basic_emotion_r}
	{r}^{j}=1, j\in \left[ 1,C \right],
	\end{align}
	indicating that only a single emotion described the specific emotional state with all its weight.
	\item \textbf{Similarity}:
	The similarity between different emotions are measured with the distance between their polar angles, \ie, image $i_1$ and image $i_2$ share the same emotion type if and only if ${\theta}_{i_1}={\theta}_{i_2}$.
	As mentioned in a psychological literature~\cite{plutchik1980general}, ``since a circle combines the concepts of degree of similarity (nearness) and degree of opposition''.
	The larger the distance between two polar angles, the greater the dissimilarities between two emotion types. 
	\item \textbf{Additivity}:
	According to psychological theories~\cite{plutchik1980general}, basic emotion vectors can be intensified and combined to produce any complex emotional states.
	Thus, each compound emotion can be defined as a weighted combination of basic emotions, which can be calculated through vector addition operations.
\end{itemize}

Based on the aforementioned attributes and properties, we develop a systematic approach to map an emotion distribution to a compound emotion vector on the Emotion Circle, as shown in Algorithm~\ref{alg:algorithm1}.
Besides the strict arithmetic procedures, we also vividly illustrate the mapping process in Fig.~\ref{fig:construction} for better comprehension.
In general, each emotion distribution is first projected on basic emotion vectors with different emotion intensities according to their description degrees.
Weighted basic emotion vectors are then combined, based on the intrinsic circular structure between emotions, to form a compound emotion vector, which can be regarded as the specific circular-structured representation of a given emotion distribution.

\begin{algorithm}[t]
	\label{alg:algorithm1}
	\caption{Circular-structured emotion representation} 
	\hspace*{0.02in} {\bf Input:}
	Emotion distribution: $\mathbf{{d}_{i}}=\left[ d_{i}^{1},...d_{i}^{C} \right] $ \\
	\hspace*{0.02in} {\bf Output:} 
	Compound emotion vector: $\mathbf{{e}_{i}} = (p_i,{\theta}_{i},r_i)$
	\begin{algorithmic}[1]
		\For{$i=1$; $i<N$; $i++$} 
		\For{$j=1$; $j<C$; $j++$} 
		\State Initialize basic emotion vectors,
		\State $\mathbf{{e}^{j}}=( {p}^{j},{\theta}^{j},{r}^{j})$;
		\State Weight basic emotion vectors,
		\State $\mathbf{e_{i}^{j}}=d_{i}^{j}\times \mathbf{{{e}^{j}}}\triangleq \left( p_{i}^{j}, \theta _{i}^{j}, r_{i}^{j}\right)$;
		\State Polar coordinate to Cartesian coordinate,
		\State $\left\{\begin{matrix}
		x_{i}^{j}=r_{i}^{j}\cos \theta _{i}^{j},    \\
		y_{i}^{j}=r_{i}^{j}\sin \theta _{i}^{j};
		\end{matrix}\right.$
		\EndFor
		\State Combination of weighted basic emotion vectors,
		\State ${{x}_{i}}=\sum_{j=1}^{C}{x_{i}^{j}},\ \ {{y}_{i}}=\sum_{j=1}^{C}{y_{i}^{j}}$;
		\State Cartesian coordinate to Polar coordinate,
		\State $\left\{\begin{matrix}
		{{r}_{i}}\!\!\!\!&=&\!\!\!\!\!\!\!\!\!\!\!\!\!\!\!\!\sqrt{x_{i}^{2}+y_{i}^{2}},    \\
		{{\theta }_{i}}\!\!\!\!&=&\!\!\!\!\arctan \left( {{y}_{i}}/{{x}_{i}} \right);
		\end{matrix}\right.$
		\If{$\theta_i \in \left[ 0,\frac{1}{2}\pi \right) \cup \left[ \frac{3}{2}\pi, 2\pi \right)$}
		\State $p_i=0$,
		\Else
		\State $p_i=1$;
		\EndIf
		\State Compound emotion vector: $\mathbf{{e}_{i}} = (p_i,{\theta}_{i},r_i)$;
		\EndFor
		\State \Return $\mathbf{{e}_{i}}$
		
	\end{algorithmic}

\end{algorithm}

\subsection{Progressive Circular Loss}
\label{sec:pc_loss}
Most of the previous methods simply implement the Kullback-Leibler (KL)~\cite{kullback1951information} loss for visual emotion distribution learning, which measure the information loss caused by the inconsistency between the predicted distribution and the labeled one:
\begin{align}
\label{eq:kl_loss}
{{\mathcal{L}}_{KL}} = -\frac{1}{N} \sum\limits_{i=1}^{N} \sum\limits_{j=1}^{C}{\mathbf{d_{i}}(j)\ln \mathbf{\hat{d}_{i}}(j),}
\end{align}
\vspace{-8pt}
\begin{align}
\label{eq:softmax}
\mathbf{\hat{d}_{i}}(j| \mathbf f_{i},\mathbf W)=\frac{\exp \left( {\mathbf{w}_{i,j}}\mathbf f_{i} \right)}{\sum_{j=1}^{C}{\exp \left( {\mathbf{w}_{i,j}} \mathbf f_{i} \right)}},
\end{align}
where $\mathbf{d_{i}}$ denotes the labeled emotion distribution from datasets and $\mathbf{\hat{d}_{i}}$ represents the predicted one.
$N$ denotes the number of images in a specific dataset and $C$ represents the involved emotion categories.


\begin{table*}
	\centering
	\scriptsize
	\caption{Comparison with the state-of-the-art methods on Flickr\_LDL dataset.}
	\label{tab:SOTA_Flickr}
	\renewcommand\arraystretch{1.00}
	\setlength{\tabcolsep}{1.3mm}{
		\begin{tabular}{ccccccccccccccc}
			\toprule
			\toprule
			&\multicolumn{2}{c}{PT}&\multicolumn{2}{c}{AA}&\multicolumn{3}{c}{SA}&\multicolumn{7}{c}{CNN-based} \\
			\cmidrule(lr){2-3}
			\cmidrule(lr){4-5}
			\cmidrule(lr){6-8}
			\cmidrule(lr){9-15}
			Measures & PT-Bayes & PT-SVM & AA-kNN & AA-BP & SA-IIS & SA-BFGS & SA-CPNN & CNNR & DLDL & ACPNN & JCDL & SSDL & E-GCN & Ours\\
			\midrule
			Chebyshev $\downarrow$ & 0.44(13) & 0.55(14) & 0.28(8) & 0.36(11) & 0.31(10) & 0.37(12) & 0.30(9) & 0.25(5) & 0.25(5) & 0.25(5) & 0.24(4) & 0.23(2) & 0.23(2) & \textbf{0.21(1)}\\
			Clark $\downarrow$ & 0.89(14) & 0.87(13) & \textbf{0.57(1)} & 0.82(8) & 0.82(8) & 0.86(12) & 0.82(8) & 0.84(11) & 0.78(5) & 0.77(2) & 0.77(2) & 0.78(5) & 0.78(5) & 0.77(2)\\
			Canberra $\downarrow$ & 0.85(14) & 0.83(13) & \textbf{0.41(1)} & 0.75(10) & 0.75(10) & 0.82(12) & 0.74(9) & 0.73(8) & 0.70(7) & 0.70(5) & 0.70(5) & 0.69(3) & 0.69(3) & 0.68(2)\\
			KL $\downarrow$ & 1.88(13) & 1.69(12) & 3.28(14) & 0.82(10) & 0.66(7) & 1.06(11) & 0.71(9) & 0.70(8) & 0.54(5) & 0.62(6) & 0.53(4) & 0.46(3) & 0.44(2) & \textbf{0.41(1)}\\
			Cosine $\uparrow$ & 0.63(13) & 0.32(14) & 0.79(7) & 0.72(9) & 0.78(8) & 0.70(11) & 0.70(11) & 0.72(9) & 0.81(5) & 0.80(6) & 0.82(4) & 0.85(3) & 0.86(2) & \textbf{0.87(1)}\\
			Intersection $\uparrow$ & 0.49(13) & 0.29(14) & 0.64(5) & 0.53(12) & 0.60(9) & 0.56(11) & 0.60(9) & 0.62(7) & 0.64(5) & 0.62(7) & 0.65(4) & 0.68(3) & 0.69(2) & \textbf{0.71(1)}\\
			\midrule
			Average Rank $\downarrow$ & 13.3(13) & 13.3(13) & 6(7) & 10(11) & 8.7(9) & 11.5(12) & 9.2(10) & 8(8) & 5.3(6) & 5.2(5) & 3.8(4) & 3.2(3) & 2.7(2) & \textbf{1.3(1)}\\
			Accuracy $\uparrow$ & 0.47(13) & 0.37(14) & 0.61(5) & 0.52(11) & 0.58(9) & 0.50(12) & 0.58(9) & 0.61(5) & 0.61(5) & 60.0(8) & 0.64(4) & 0.70(2) & 0.69(3) & \textbf{0.72(1)}\\
			\bottomrule
			\bottomrule
	\end{tabular}}
	\vspace{-5pt}
\end{table*}

\begin{table*}
	\centering
	\scriptsize
	\caption{Comparison with the state-of-the-art methods on Twitter\_LDL dataset.}
	\label{tab:SOTA_Twitter}
	\renewcommand\arraystretch{1.00}
	\setlength{\tabcolsep}{1.3mm}{
		\begin{tabular}{ccccccccccccccc}
			\toprule
			\toprule
			&\multicolumn{2}{c}{PT}&\multicolumn{2}{c}{AA}&\multicolumn{3}{c}{SA}&\multicolumn{7}{c}{CNN-based} \\
			\cmidrule(lr){2-3}
			\cmidrule(lr){4-5}
			\cmidrule(lr){6-8}
			\cmidrule(lr){9-15}
			Measures & PT-Bayes & PT-SVM & AA-kNN & AA-BP & SA-IIS & SA-BFGS & SA-CPNN & CNNR & DLDL & ACPNN & JCDL & SSDL & E-GCN & Ours\\
			\midrule
			Chebyshev $\downarrow$ & 0.53(13) & 0.63(14) & 0.28(7) & 0.37(11) & 0.28(7) & 0.37(11) & 0.36(10) & 0.28(7) & 0.26(5) & 0.27(6) & 0.25(3) & 0.25(3) & 0.24(2) & \textbf{0.22(1)}\\
			Clark $\downarrow$ & 0.85(7) & 0.91(14) & \textbf{0.58(1)} & 0.89(12) & 0.86(11) & 0.89(12) & 0.85(7) & 0.84(3) & 0.84(3) & 0.85(7) & 0.83(2) & 0.84(3) & 0.85(7) & 0.84(3)\\
			Canberra $\downarrow$ & 0.77(6) & 0.88(14) & \textbf{0.41(1)} & 0.84(12) & 0.79(11) & 0.84(12) & 0.78(8) & 0.76(2) & 0.77(6) & 0.78(8) & 0.76(2) & 0.76(2) & 0.78(8) & 0.76(2)\\
			KL $\downarrow$ & 1.31(12) & 1.65(13) & 3.89(14) & 1.19(10) & 0.64(7) & 1.19(10) & 0.85(9) & 0.67(7) & 0.54(5) & 0.58(6) & 0.53(4) & 0.51(3) & 0.46(2) & \textbf{0.44(1)}\\
			Cosine $\uparrow$ & 0.53(13) & 0.25(14) & 0.82(7) & 0.71(11) & 0.82(7) & 0.71(11) & 0.75(10) & 0.82(7) & 0.83(6) & 0.84(5) & 0.85(4) & 0.86(3) & 0.87(2) & \textbf{0.89(1)}\\
			Intersection $\uparrow$ & 0.40(13) & 0.21(14) & 0.66(5) & 0.59(9) & 0.63(8) & 0.57(11) & 0.56(12) & 0.58(10) & 0.65(6) & 0.64(7) & 0.68(4) & 0.69(3) & 0.70(2) & \textbf{0.72(1)}\\
			\midrule
			Average Rank $\downarrow$ & 10.7(12) & 13.8(14) & 5.8(6) & 10.8(11) & 8.5(9) & 11.2(13) & 9.3(10) & 6(7) & 5.2(5) & 6.5(8) & 3.2(3) & 2.8(2) & 3.8(4) & \textbf{1.5(1)}\\
			Accuracy $\uparrow$ & 0.45(13) & 0.40(14) & 0.73(7) & 0.72(9) & 0.70(10) & 0.57(12) & 0.70(10) & 0.74(5) & 0.73(7) & 0.74(5) & 0.76(3) & 0.77(2) & 0.76(3) & \textbf{0.78(1)}\\
			\bottomrule
			\bottomrule
	\end{tabular}}
	\vspace{-10pt}
\end{table*}

However, rather than a set of uncorrelated labels, emotions are intrinsically distributed in a circular structure in psychological models~\cite{mikels2005emotional, zhao2016predicting}. 
In order to effectively exploit such prior knowledge, we propose a Progressive Circular (PC) loss to learn the emotion distribution from coarse to fine.
The proposed PC loss, conducted on the Emotion Circle described in Sec.~\ref{sec:emotion_circle}, is designed to penalize the difference between two emotion vectors, \ie, the labeled one $\mathbf{{e}_{i}} = (p_i, \theta_i, r_i)$ and the predicted one $\mathbf{\hat{e}_{i}} = (\hat{p}_i, \hat{\theta}_i, \hat{r}_i)$.
To be specific, we progressively establish constraints on three attributes of emotion vectors, \ie, emotion polarity ($p_i$), emotion type ($\theta_i$) and emotion intensity ($r_i$).
Since polarity is the coarse attribute of an emotion, we first devise polar loss to measure the differences between emotion polarities, by implementing the Mean Square Error (MSE) loss:

\begin{align}
\label{eq:polar_loss}
{{\mathcal{L}}_{p}}=\frac{1}{N}\sum\limits_{i=1}^{N}(p_{i} -\hat{p}_{i})^{2},
\end{align}
where $p_{i}$ denotes the labeled emotion polarity and $\hat{p}_{i}$ represents the predicted one.
Eq.~\ref{eq:polar_loss} ensures the correctness in emotion polarity prediction, which is the first goal towards accurate emotion prediction. 
As mentioned in Sec.~\ref{sec:emotion_circle}, distance between polar angles can measure the similarities between different emotion types.
Thus, after divide emotions into two polarities, we build up a more fine-grained constraint on emotion type, denoted as type loss:
\begin{align}
\label{eq:type_loss}
{{\mathcal{L}}_{t}}=\frac{1}{N}\sum\limits_{i=1}^{N}(\theta_{i}-\hat{\theta}_{i})^{2}.
\end{align}

In Eq.~\ref{eq:type_loss}, the closer the two polar angles are, the more similarity lies in their emotional states.
However, it is insufficient to merely take emotion type into account, as emotion intensity is also considered as a crucial factor towards a specific emotional state~\cite{plutchik1980general}.
Suppose there are two images with the same emotion type, \ie, $\theta_{i_1}=\theta_{i_2}$, and quite different emotion intensities, \ie, $r_{i_1}=1$, $r_{i_2}=0.01$, it is really hard to group them into the same emotional state.
Therefore, serving as the confidence degree for both emotion type $\theta_i$ and emotion polarity $p_i$, we further add emotion intensity $r_i$ to the proposed loss function for a precise and detailed constraint for visual emotion distribution learning: 
\begin{align}
\label{eq:pc_loss}
{{\mathcal{L}}_{PC}}=\frac{1}{N}\sum\limits_{i=1}^{N}r_{i}\left((p_{i} -\hat{p}_{i})^{2}+(\theta_{i}-\hat{\theta}_{i})^{2}\right).
\end{align}

Based on the Emotion Circle, our PC loss is eventually constructed with three constraints in a coarse-to-fine manner in Eq.~\ref{eq:pc_loss}.
Our loss function is integrated with both KL loss and PC loss in a weighted combination:
\begin{align}
\label{eq:loss}
{{\mathcal{L}}}= (1-\mu){\mathcal{L}}_{KL} + \mu {\mathcal{L}}_{PC},
\end{align}
where $\mu$ is a hyper-parameter balancing the importance between the two losses and is further ablated in Sec.~\ref{sec:hyperparameter_analysis}.
So far, the emotion distribution is not only learned in a conventional mechanism, but further gains a performance boost in an emotion-specific and circular-structured manner.

\section{Experimental Results}
\label{sec:experimental_results}
\subsection{Datasets}
\label{sec:datasets}
We evaluate our method on three public visual emotion distribution datasets, including Flickr\_LDL, Twitter\_LDL~\cite{yang2017learning} and Abstract Paintings~\cite{machajdik2010affective}.
Built on Mikel's eight emotion space, there are 11,150 images in Flickr\_LDL and 10,045 images in Twitter\_LDL.
Eleven viewers are hired to label Flickr\_LDL and eight viewers are hired to label Twitter\_LDL, where the detailed votes from all the workers are integrated to generate the ground truth label distribution for each image.
The Abstract Paintings dataset is also labeled with Mikel's eight emotions, which consists only of combinations of color and texture without any recognizable objects and contains 280 images in total.

\begin{table*}
	\centering
	\scriptsize
	\caption{Comparison with the state-of-the-art methods on Abstract Paintings dataset.}
	\label{tab:SOTA_Abstarct}
	\renewcommand\arraystretch{1.00}
	\setlength{\tabcolsep}{3mm}{
	\begin{tabular}{ccccccccc}
		\toprule
		\toprule
		&\multicolumn{2}{c}{PT}&\multicolumn{2}{c}{AA}&\multicolumn{2}{c}{SA}&\multicolumn{2}{c}{CNN-based} \\
		\cmidrule(lr){2-3}
		\cmidrule(lr){4-5}
		\cmidrule(lr){6-7}
		\cmidrule(lr){8-9}
		Measures & PT-Bayes & PT-SVM & AA-kNN & AA-BP & SA-IIS & SA-BFGS & ACPNN & Ours\\
		\midrule
		Chebyshev $\downarrow$ &0.360(7)&0.298(6)&0.245(3)&0.297(5)&0.296(4)&0.472(8)&0.234(2)&\textbf{0.226(1)}\\
		Clark $\downarrow$ &0.674(7)&0.632(4)&0.621(2)&0.641(5)&0.660(6)&0.834(8)&0.625(3)&\textbf{0.562(1)}\\
		Canberra $\downarrow$ &0.594(7)&0.537(4)&0.513(2)&0.544(6)&0.564(5)&0.781(8)&0.522(3)&\textbf{0.456(1)}\\
		KL $\downarrow$ &3.268(8)&0.708(5)&0.515(3)&0.782(6)&0.644(4)&2.282(7)&0.513(2)&\textbf{0.441(1)}\\
		Cosine $\uparrow$ &0.653(5)&0.643(6)&0.753(3)&0.636(7)&0.692(4)&0.573(8)&0.763(2)&\textbf{0.773(1)}\\
		Intersection $\uparrow$ &0.518(7)&0.539(6)&0.608(3)&0.540(5)&0.577(4)&0.417(8)&0.618(2)&\textbf{0.654(1)}\\
		\midrule
		Average Rank $\downarrow$ &6.83(7)&5.17(6)&2.68(3)&4.83(5)&3.83(4)&7.83(8)&2.33(2)&\textbf{1(1)}\\
		\bottomrule
		\bottomrule
	\end{tabular}}
	\vspace{-10pt}
\end{table*}

\begin{table}
	\centering
	\scriptsize
	\caption{Ablation study of loss function on Flickr\_LDL dataset.}
	\label{tab:ablation_Flickr}
	\renewcommand\arraystretch{1}
	\setlength{\tabcolsep}{0.8mm}{
		\begin{tabular}{cp{1cm}<{\centering}p{1cm}<{\centering}p{1.2cm}<{\centering}p{1.5cm}<{\centering}p{1.2cm}<{\centering}p{1cm}}
			\toprule
			\toprule
			Measures & ${\mathcal{L}}_{K\!L}$ & ${\mathcal{L}}_{K\!L}\!\!+\!\!{\mathcal{L}}_{p}$ &${\mathcal{L}}_{K\!L}\!\!+\!\!{\mathcal{L}}_{t}$ & ${\mathcal{L}}_{K\!L}\!+\!{\mathcal{L}}_{p}\!+\!{\mathcal{L}}_{t}$ & ${\mathcal{L}}_{K\!L}\!\!+\!\!{\mathcal{L}}_{P\!C}$\\
			\midrule
			Chebyshev $\downarrow$ & 0.239 & 0.225 & 0.222 & 0.218 & \textbf{0.213} \\
			Clark $\downarrow$ & 0.783 & 0.779 & 0.779 & 0.775 & \textbf{0.774} \\
			Canberra $\downarrow$ & 0.697 & 0.689 & 0.687 & \textbf{0.682} & 0.685\\
			KL $\downarrow$ & 0.435 & 0.441 & 0.420 & 0.414 & \textbf{0.408}\\
			Cosine $\uparrow$ & 0.843 & 0.862 & 0.869 & 0.870 & \textbf{0.874}\\
			Intersection $\uparrow$ & 0.678 & 0.693 & 0.705 & 0.703 & \textbf{0.709} \\
			Accuracy $\uparrow$ & 0.669 & 0.695 & 0.700 & 0.718 & \textbf{0.721}\\
			\bottomrule			\bottomrule	\end{tabular}}
	\vspace{-5pt}
\end{table}

\begin{table}
	\centering
	\scriptsize
	\caption{Ablation study of loss function on Twitter\_LDL dataset.}
	\label{tab:ablation_Twitter}
	\renewcommand\arraystretch{1}
	\setlength{\tabcolsep}{0.8mm}{
		\begin{tabular}{cp{1cm}<{\centering}p{1cm}<{\centering}p{1.2cm}<{\centering}p{1.5cm}<{\centering}p{1.2cm}<{\centering}p{1cm}}
			\toprule
			\toprule
			Measures & ${\mathcal{L}}_{K\!L}$ & ${\mathcal{L}}_{K\!L}\!\!+\!\!{\mathcal{L}}_{p}$ &${\mathcal{L}}_{K\!L}\!\!+\!\!{\mathcal{L}}_{t}$ & ${\mathcal{L}}_{K\!L}\!+\!{\mathcal{L}}_{p}\!+\!{\mathcal{L}}_{t}$ & ${\mathcal{L}}_{K\!L}\!\!+\!\!{\mathcal{L}}_{P\!C}$\\
			\midrule
			Chebyshev $\downarrow$ & 0.259 & 0.240 & 0.233 & 0.230 & \textbf{0.224} \\
			Clark $\downarrow$ & 0.861 & 0.851 & 0.848 & 0.846 & \textbf{0.842}\\
			Canberra $\downarrow$ & 0.797 & 0.778 & 0.775 & 0.772 & \textbf{0.764}\\
			KL $\downarrow$  & 0.464 & 0.476 & 0.455 & 0.450 & \textbf{0.439}\\
			Cosine $\uparrow$  & 0.848 & 0.870 & 0.878 & 0.882 & \textbf{0.886}\\
			Intersection $\uparrow$  & 0.686 & 0.706 & 0.703 & 0.713 & \textbf{0.717}\\
			Accuracy $\uparrow$  & 0.744 & 0.764 & 0.770 & 0.779 & \textbf{0.781}\\
			\bottomrule
			\bottomrule
	\end{tabular}}
	\vspace{-10pt}
\end{table}

\subsection{Implementation Details}
\label{sec:implementation_details]}·
In the experiment, our backbone network is built on ResNet-50~\cite{he2016deep} following the same setting as previous methods~\cite{he2019image}, which is pre-trained on a large-scale visual recognition dataset, ImageNet~\cite{deng2009imagenet}.
Specifically, the output dimension of the last fully connected layer is changed to emotion numbers according to the datasets. 
Following the same setting in~\cite{yang2017learning}, Flickr\_LDL, Twitter\_LDL and  Abstract Paintings datasets are randomly split into training set (80\%) and testing set (20\%). 
For training/testing sets, after resizing each image to 480 on its shorter side, we then crop it to 448$\times$448 randomly followed by a horizontal flip~\cite{he2016deep}.
The whole network is trained by the adaptive optimizer Adam~\cite{kingma2014adam} in an end-to-end manner with KL loss and the proposed PC loss.
With a weight decay of 5e-5, the learning rate of Adam starts from 1e-5 and divided by 10 every 10 epochs, and the total epoch number is set to 50.
Our framework is implemented using PyTorch~\cite{paszke2017automatic} and our experiments are performed on an NVIDIA GTX 1080Ti GPU.

\subsection{Comparison with the State-of-the-art Methods}
\label{sec:sota}
To evaluate the effectiveness of the proposed circular-structured representation, we conduct extensive experiments compared with the state-of-the-art methods on three public visual emotion distribution datasets, including Flickr\_LDL, Twitter\_LDL and Abstract Paintings, as shown in TABLE~\ref{tab:SOTA_Flickr}, TABLE~\ref{tab:SOTA_Twitter} and TABLE~\ref{tab:SOTA_Abstarct} respectively.
In general, the state-of-the-art methods can be divided into four types: PT, AA, SA and CNN-based.

\begin{itemize}
	\setlength{\itemsep}{0pt}
	\setlength{\parsep}{0pt}
	\setlength{\parskip}{0pt}
	\item \textbf{Problem Transformation (PT}):
	Based on the representative algorithms Bayes classifier and SVM, PT-Bayes and PT-SVM are designed to transfer the LDL problem into a single-label learning (SLL) one~\cite{geng2016label}.
	Since these methods roughly turn a complex LDL task into a simple SLL one, performance drops may exist when measuring the differences between distributions.
	
	\item \textbf{Algorithm Adaptation (AA)}:
	The existing machine learning algorithms k-NN and BP neural network are extended to deal with label distributions, which are denoted by AA-kNN and AA-BP respectively~\cite{geng2016label}.
	In TABLE~\ref{tab:SOTA_Flickr} and TABLE~\ref{tab:SOTA_Twitter}, AA-kNN achieves insurmountable results in Clark distance and Canberra metric, owing to its superiority in dealing with overlapping samples in visual emotion distributions.
	
	\item \textbf{Specialized Algorithm (SA}): 
	Specialized algorithms are designed by directly matching the characteristics of LDL, including SA-IIS, SA-BFGS, SA-CPNN~\cite{geng2016label, geng2013facial}.
	By implementing a similar strategy to Improved Iterative Scaling (IIS), SA-IIS assumes the parametric model to be the maximum entropy mode~\cite{geng2016label, geng2013facial}.
	However, IIS often performs worse compared with other optimization algorithms such as conjugate gradient and quasi-Newton methods.
	Thus, SA-BFGS is further developed with an effective quasi-Newton method named Broyden-Fletcher-Goldfarb-Shanno for LDL~\cite{geng2016label}. 
	SA-CPNN is proposed based on a three-layer conditional probability neural network~\cite{geng2013facial}.
	Compared with the above PT and AA methods, SA has achieved improved results for their specially designed features.
	
	\item \textbf{CNN-based methods (CNN-based)}:
	Benefiting from its powerful representation ability, CNN-based methods gain a significant performance boost compared with those traditional ones.
	Specifically, CNNR~\cite{peng2015mixed} is proposed to treat VEA in a label distribution paradigm for the first time, which achieved a great performance with an end-to-end network.
	DLDL~\cite{gao2017deep} replaces the Euclidean loss with KL loss to well describe the differences between two distributions.
	In order to learn visual emotion distribution and classification in a joint manner, JCDL~\cite{yang2017joint} further boosts the performance by optimizing both KL loss and softmax loss. 
	
\end{itemize}

\begin{figure}
	\centering
	\includegraphics[width=0.98\linewidth]{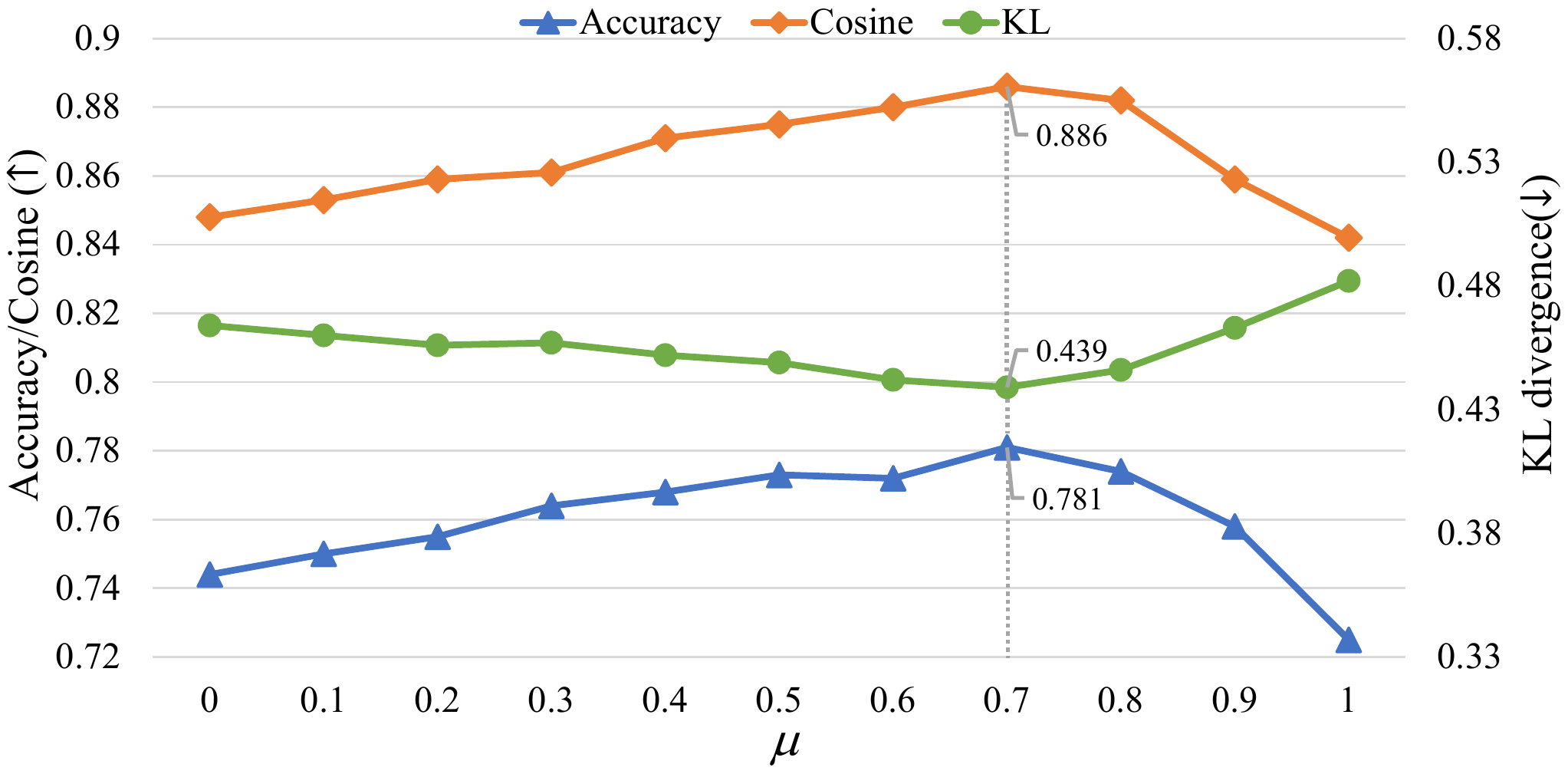}
	\vspace{-5pt}
	\caption{Effect of $\mu$ for combined loss on Twitter\_LDL dataset.
		Note that $\mu = 1$ suggests only using PC loss while
		$\mu = 0$ means implementing KL loss alone.
	}
	\vspace{-10pt}
	\label{fig:ablation}
\end{figure}

\begin{figure*}
	\centering
	\includegraphics[width=0.98\textwidth]{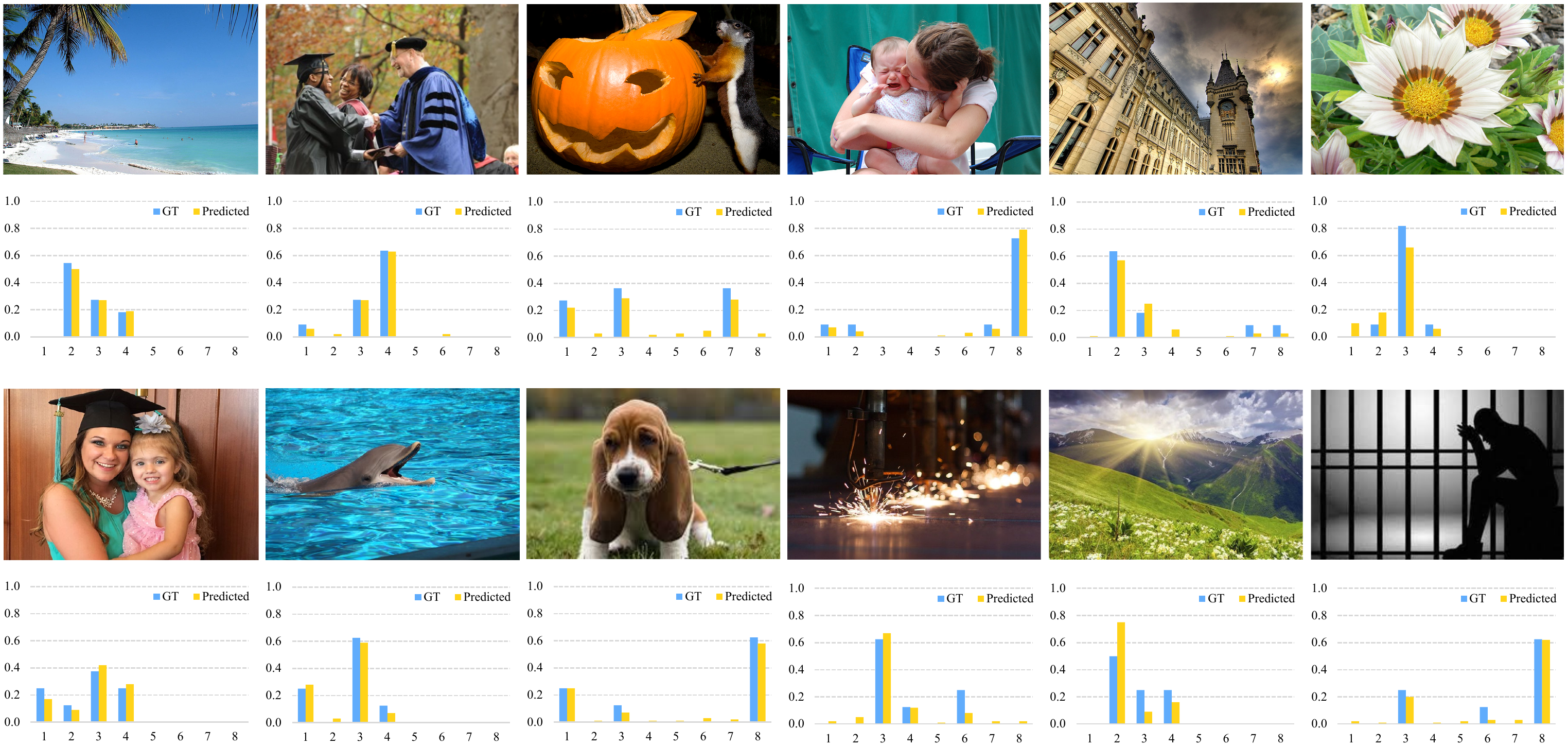}
	\vspace{-0pt}
	\caption{Visualization of the predicted emotion distributions (predicted) and the ground-truth (GT) ones, where images in the first line come from Flickr\_LDL dataset and second line the Twitter\_LDL. Each number on the horizontal axis corresponds to an emotion category. 
	}
	\vspace{-10pt}
	\label{fig:visual}
\end{figure*}

Following the same setting as the previous work~\cite{geng2016label, he2019image, xiong2019structured, yang2017joint, yang2017learning}, we evaluate the performance of visual emotion distribution learning task with six commonly-used measurements, \ie, Chebyshev distance ($\downarrow$), Clark distance ($\downarrow$), Canberra metric ($\downarrow$), Kullback-Leibler divergence ($\downarrow$), cosine coefficient ($\uparrow$), and intersection similarity ($\uparrow$).
Among these measurements, the first four are distance measures and the last two are similarity measures, where $\downarrow$ suggests the lower the better and $\uparrow$ the higher the better.
As KL divergence is not well defined with zero value, we use a small value $\varepsilon ={{10}^{-10}}$ to approximate it.
Besides, since the maximum values of Clark distance and Canberra metric are determined by the number of emotions, we divided Clark distance by the square root of number of emotions and divided Canberra metric by the number of emotions for standardized comparisons.
In addition to the above six measurements, we further introduce top-1 accuracy as another evaluation metric, which ensures the prediction results of the dominant emotions in the distributions. 
From the above analysis, it is obvious that the proposed method consistently outperforms the state-of-the-art methods on three widely-used datasets, owing to the effectiveness in circular-structured representation for visual emotion distribution learning.
 
\subsection{Ablation Study}
\label{sec:ablation_study}
\subsubsection{Effectiveness of the PC Loss}
\label{sec:effectiveness_pc_loss}
In TABLE~\ref{tab:ablation_Flickr} and TABLE~\ref{tab:ablation_Twitter}, we conduct ablation study on the proposed PC loss with two large-scale datasets, aiming to verify the effectiveness of each proposed constraint (\ie, polar loss, type loss and emotion intensity).
By adopting KL loss alone, we first evaluate the baseline of our method with six measurements and top-1 accuracy.
Polarity loss and type loss are then separately and jointly added to KL loss, where both bring significant performance boosts as emotion polarity and emotion type serve as the basic and decisive attributes in visual emotion distribution learning.
For a more fine-grained description, emotion intensity further improves all the measurements and gains the optimal result eventually.
From the above analysis, we can conclude that each constraint in PC loss is indispensable, jointly and progressively contributing to the final result.

\subsubsection{Hyper-Parameter Analysis}
\label{sec:hyperparameter_analysis}
As $\mu$ controls the relative importance between the proposed PC loss and KL loss, we conduct experiments to validate the choice of $\mu = 0.7$ in Eq.~\ref{eq:loss}, as shown in Fig.~\ref{fig:ablation}.
The greater the value of $\mu$, the more importance lies in the proposed PC loss.
Based on the Twitter\_LDL dataset, we implement three measurements, namely KL divergence, Cosine coefficient and Accuracy, to demonstrate how $\mu$ influences the performance of the proposed method.
We find that all the three measurements constantly grows as $\mu$ varies from 0 to 0.7 and drops after $\mu = 0.7$, which suggests that a combination of PC loss and KL loss achieves the best performance.
As KL loss solves the problem of general label distribution learning while PC loss considers emotion-specific circular structure as prior knowledge, each loss encounters performance drop when acting alone, which further suggests that both losses are essential and complementary in visual emotion distribution learning.  

\subsection{Visualization}
\label{sec:visualization}
\vspace{-3pt}

We further visualize the predicted emotion distributions (predicted) and the ground-truths (GT) on both Flickr\_LDL and Twitter\_LDL datasets in Fig.~\ref{fig:visual}.
It is obvious that the proposed method restores the original emotion distribution to a large extent on both emotion polarity, emotion type and emotion intensity.
Specifically, we can infer from the visualized results that once an emotion distribution only involves a single polarity, the prediction is more accurate with the help of the polar loss.
When encounters complex emotional state, \ie, an emotional state is made up of different polarized emotions, our proposed method can still achieve relatively good results, owing to the effectiveness of the proposed type loss and emotion intensity.

\vspace{-5pt}
\section{Conclusion}
\label{sec:conclusion}
\vspace{-3pt}

In this paper, we have proposed a circular-structured representation for visual emotion distribution learning by exploiting the intrinsic relationships between emotions based on psychological models.
We first constructed the Emotion Circle to present any emotional state with emotion vectors and then designed a Progressive Circular (PC) loss to constraint emotion vectors in a coarse-to-fine manner.
Extensive experiments and comparisons have shown that the proposed method consistently outperforms the state-of-the-art methods on three visual emotion distribution datasets.

\noindent\textbf{Acknowledgments:}
This work was supported in part by the National Natural Science Foundation of China under Grants (62036007, 61772402, 62050175, 61972305, 61871308 and 61771473), Natural Science Foundation of Jiangsu Province under Grant (BK20181354).


{\small
\bibliographystyle{ieee_fullname}
\bibliography{egbib}
}

\end{document}